\documentclass[letterpaper, 10 pt, conference]{ieeeconf}  % Comment this line out
\IEEEoverridecommandlockouts                              % This command is only
\usepackage{caption}
\usepackage{esvect}
\usepackage[table]{xcolor}
\usepackage{graphicx}
\usepackage{subfig}
\usepackage{multirow}
\usepackage{hyperref}
\hypersetup{
    colorlinks=false,
    linkcolor=blue,
    filecolor=magenta,      
    urlcolor=cyan,
}
\usepackage{mathptmx} 
\usepackage{times} 
\usepackage{amssymb}
\usepackage{mathtools}

\usepackage{bm}
\usepackage{ragged2e}
\usepackage{ragged2e}

\title{\LARGE \bf
A Supervised Learning Methodology for Real-Time Disguised Face Recognition in the Wild
}

\author{Saumya Kumaar$^{3}$, Abhinandan Dogra$^{4}$, Abrar Majeedi$^{4}$, Hanan Gani$^{4}$, Ravi M. Vishwanath$^{2}$ and S N Omkar$^{1}$
\thanks{$^{1}$ Chief Research Scientist, Indian Institute of Science, Bangalore}
\thanks{$^{2}$ Research Associate, Indian Institute of Science, Bangalore}
\thanks{$^{3}$ Research Assistant, Indian Institute of Science, Bangalore}
\thanks{$^{4}$ Student, National Institute of Technology, Srinagar}
}
\begin{document}

\maketitle
\thispagestyle{empty}
\pagestyle{empty}

%%%%%%%%%%%%%%%%%%%%%%%%%%%%%%%%%%%%%%%%%%%%%%%%%%%%%%%%%%%%%%%%%%%%%%%%%%%%%%%%
\begin{abstract}

Facial recognition has always been a challenging task for computer vision scientists and experts. Despite complexities arising due to variations in camera parameters, illumination and face orientations, significant progress has been made in the field with deep learning algorithms now competing with human-level accuracy. But in contrast to the recent advances in face recognition techniques, Disguised Facial Identification continues to be a tougher challenge in the field of computer vision. The modern day scenario, where security is of prime concern, regular face identification  techniques do not perform as required when the faces are disguised, which calls for a different approach to handle situations where intruders have their faces masked. Along the same lines, we propose a deep learning architecture for disguised facial recognition (DFR). The algorithm put forward in this paper detects 20 facial key-points in the first stage, using a 14-layered convolutional neural network (CNN). These facial key-points are later utilized by a support vector machine (SVM) for classifying the disguised faces based on the euclidean distance ratios and angles between different facial key-points. This overall architecture imparts a basic intelligence to our system. Our key-point feature prediction accuracy is 65\% while the classification rate is 72.4\%. Moreover, the architecture works at 19 FPS, thereby performing in almost real-time. The efficiency of our approach is also compared with the state-of-the-art Disguised Facial Identification methods.

\end{abstract}

%%%%%%%%%%%%%%%%%%%%%%%%%%%%%%%%%%%%%%%%%%%%%%%%%%%%%%%%%%%%%%%%%%%%%%%%%%%%%%%%
\section{Introduction}
The domain of face verification, or what is commonly referred to as facial recognition nowadays, has witnessed significant advances in the past few decades. Facial recognition is a classic example of a modern real-world challenge and it is directly implemented in places that are required to be impregnable to any kind of human intrusion.

The concept has been tested and implemented widely and has laid the foundation for several other feature-extraction based algorithms for general object detection. One of the most commonly used algorithm for face description and disambiguation is the \textit{eigenfaces} architecture, as proposed by Turk \textit{et. al.} [1], which transformed the approach to tackling face verification. Seven years down the line elastic bunch graph matching [2] and virtual eigensignatures [3]
became quite popular as standard face description techniques. Later in the year 2006, Ahonen \textit{et. al.} [4] came up with a novel technique called \textit{local binary patterns}, which became one of the most extensively used face descriptors of all times. Even OpenCV's ([22] and [23]) default descriptor function for faces is also based on the same algorithm, and is called LBPFaceRecognizer. Interestingly, around the same time as [2] and [3], neural networks were gaining popularity in the field of machine learning. People have implemented decision-based [5] neural nets and convolutional neural nets [6] for face recognition tasks, and with the recent invention of the concept of light convolutional neural network by Wu \textit{et. al.} [7], face recognition has now seemed to have reached the zenith of facial classification rates. On the other hand, due to the complexities present in disguised face identification, it has not witnessed much research focus in the past few decades.

In the modern world, tons of cases have been reported where masked intruders have been accused of several wrongdoings. Nonetheless, identification of such people becomes a problem when the autonomous facial recognition systems fail to perform. There are security measures almost everywhere on this planet but very few have the capabilities of isolating and identifying disguised faces in the wild, and the need for such a development is on the rise.  But, due to inherent complexities of the problem, previous research results on the same, indicate more of chance-based performance rather than a feature-based classification. 

An interesting approach to solving problems related to facial expression detection or face recognition itself, is by extracting facial key-points. For instance, Berretti \textit{et. al.} [8] have used SIFT descriptors of auto-detected keypoints for examining facial expressions. They trained a SVM for every single facial expression and were able to achieve a classification rate of 78.43\% on the BU-3DFE database. Sun \textit{et. al.} [9] attempted to predict five facial key-points, the left eye center, right eye center, nose tip, left mouth corner and right mouth corner and they cascaded three levels of convolutional networks to perform a \textit{coarse-to-fine} tuning of the predictions. Wang \textit{et. al.} [10] have used histogram stretching and principal component analysis on the stretched images to obtain the eigenfaces. In these so obtained eigenfaces they performed their mean patch searching algorithm to predict the left and right eye centres, for any input face image. Very recently, Shi \textit{et. al.} [11] used principal component analysis and local binary patterns descriptor to process the data and they reviewed the outputs of various algorithms to that processed data like linear regression, tree based model, CNNs etc. for facial key-point detection. The paper provided a lot of insights to the various architectures for the same. 

However, to our knowledge, using key-points for identifying disguised faces was a method which was not adopted and people continued to use the existing face descriptors for the same. Yoshine \textit{et. al.} [12] had suggested a method of morphometrical matching for identifying disguised faces, where they took the 2D right oblique images (with three disguises only) and superimposed them on 100 different subjects. They reported a difference of 2.3 - 2.8mm for a match (which approximately comes to an offset of 11 pixels). Singh \textit{et. al.} [13] suggested an algorithm called 2DLPGNN which  resulted in the best verification performance. But for the challenging scenarios of multi-disguise variations, the accuracy drops to 65.6\%  This research however proved that existing algorithms are not effective enough to handle a substantial degree of disguise variation. Righi \textit{et. al.} [14] conducted three different visual cognition experiments to observe the effects of disguises on observer's face recognition activity using natural images in which individuals had a mixture of eyeglasses and wigs as disguises. Their paper presented some fundamental cognitive techniques that humans use to disambiguate disguised faces. But the method suggested by Dhamecha \textit{et. al.} [15] proved to be state-of-the-art classification technique for disguised faces, when they reported an accuracy of 53\%. Facial Key points based disguised face verification was first used in 2017 by Singh \textit{et. al.} [16] where they proposed to use an existing architecture of spatial fusion convolutional network, which was suggested earlier by Pfister \textit{et. al.} [20] for human pose estimation. They were however, able to significantly improve the previous state-of-the-art performance by a margin of 9\%. 

Inspired by the above mentioned works, we present a novel solution for disguised face recognition, the DFR model. We propose to use a CNN for key-points prediction and a support vector machine (SVM) for classification. It is a common understanding that SVMs are meant for real-time applications, and that is the precise motivation behind using it. The primary contributions of this paper could be listed down as :

\begin{itemize}
\item A novel CNN architecture for enhanced key-point feature detection coupled with a classification technique that attempts to give a real-time performance at 19 Hz.

\item The architecture so presented, has been made user-friendly in terms of deployment to real-world problems like intruder detection etc.
\end{itemize}

\section{Methodology}
The overall system architecture is depicted in Fig. 1. The sub-sections under methodology are divided accordingly.

\begin{figure}[h]
\centering
\includegraphics[width=3.5 in]{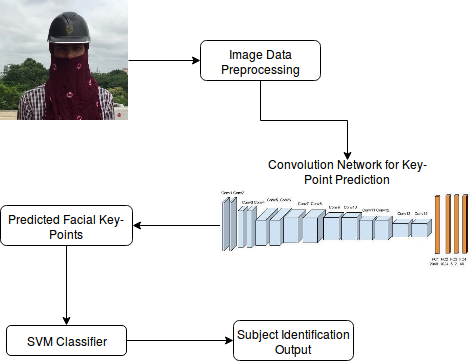}
\caption{Overall System Architecture}
\label{fig_chi_dot}
\end{figure}

\subsection{Dataset}

\begin{figure}[h]
\centering
\includegraphics[width=3.2 in]{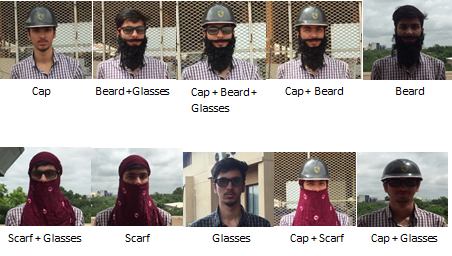}
\caption{Different Disguises Used in the Experiments}
\label{fig_chi_dot}
\end{figure}

For any CNN to work effectively, there should be enough amount of training data available. So in order to benchmark our architecture, we have used the same dataset as proposed by Singh \textit{et. al.} [16]. So in most of the literature survey, we observed that even though the problem of DFR was tackled with reasonable accuracies, the datasets consisted of purely faces with minimal backgrounds. We are of the opinion that such datasets might cause the models to exhibit a degraded performance in real life scenarios where there is a lot of background noise. It is worth mentioning here that the faces have not been cropped and the backgrounds has not been altered. So that allows us to have a mixture of simple, semi-complex and complex backgrounds in our dataset as shown in Fig. 2.

\begin{figure}[h]
\centering
\includegraphics[width=3.2 in]{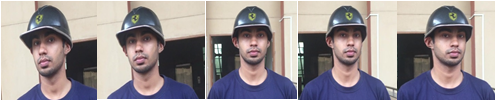}
\caption{Different Orientations of Disguised Faces}
\label{fig_chi_dot}
\end{figure}

\begin{figure}
    \centering
    \subfloat{{\includegraphics[width=3cm]{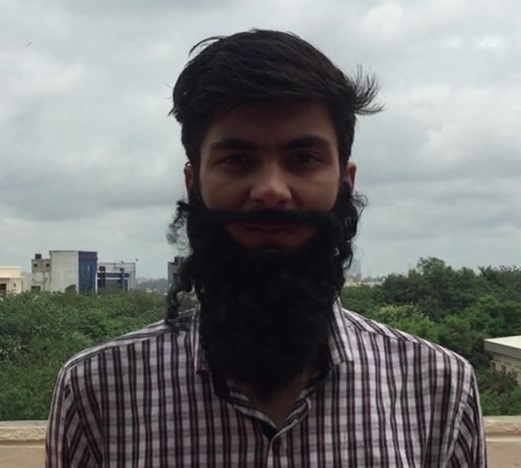} }}
    \subfloat{{\includegraphics[width=3cm]{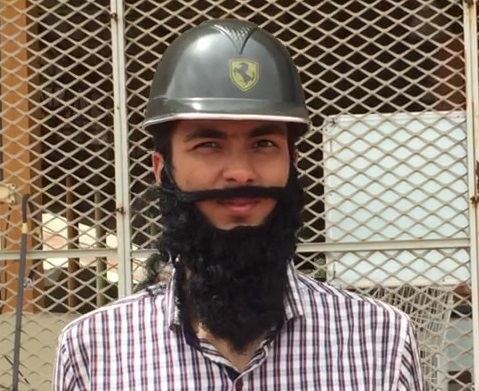} }}
    \quad
        \subfloat{{\includegraphics[width=3cm]{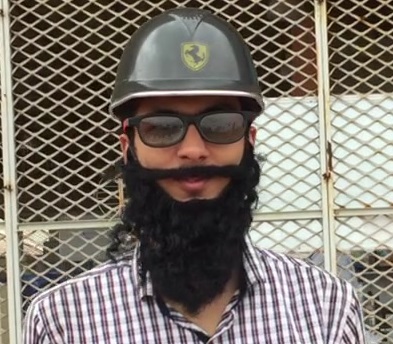} }}
    \subfloat{{\includegraphics[width=3cm]{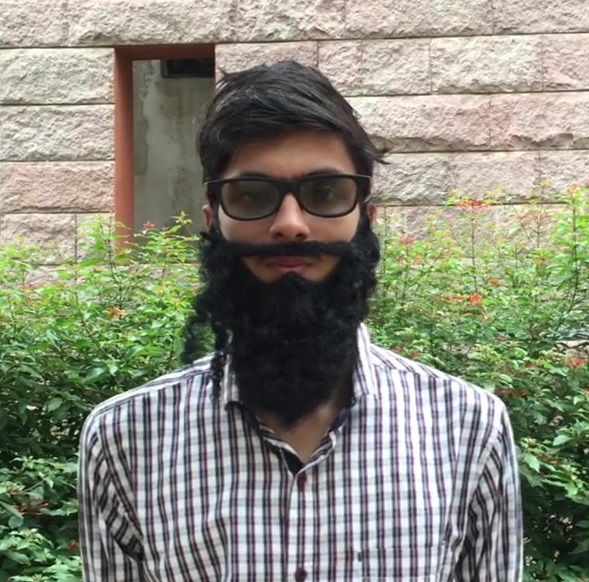} }}
            \quad
        \subfloat{{\includegraphics[width=3cm]{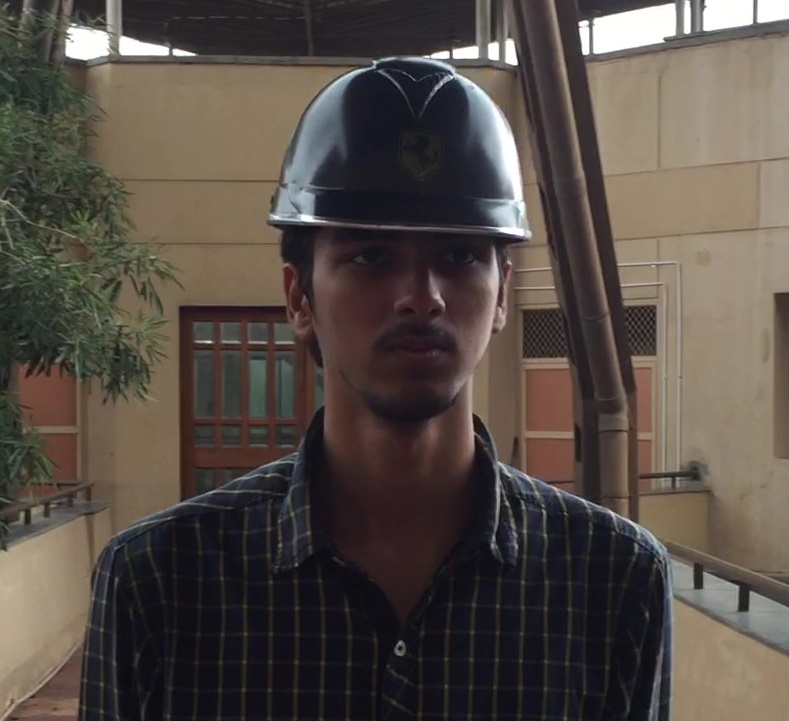} }}
    \subfloat{{\includegraphics[width=3cm]{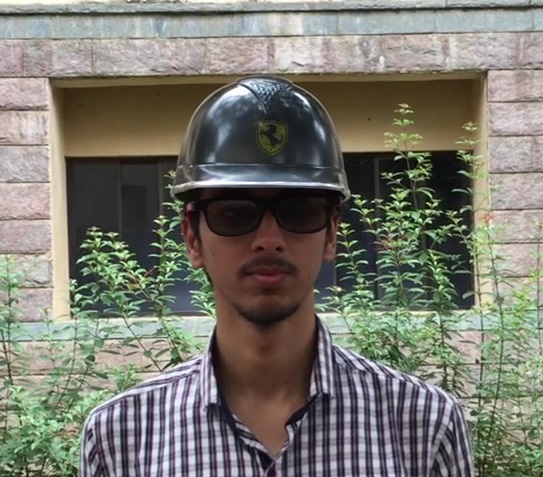} }}
    \quad
        \subfloat{{\includegraphics[width=3cm]{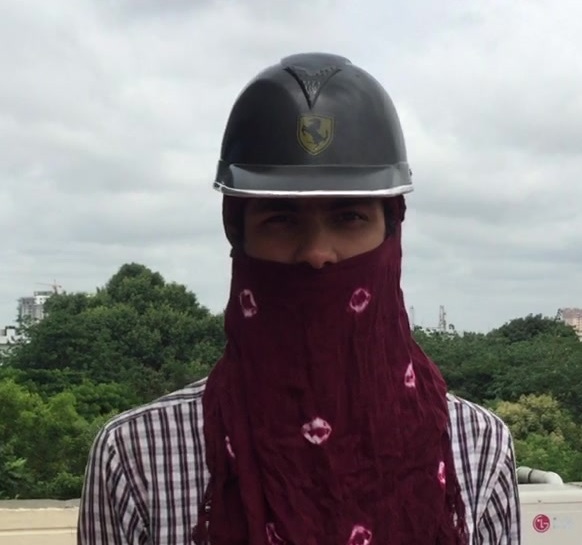} }}
    \subfloat{{\includegraphics[width=3cm]{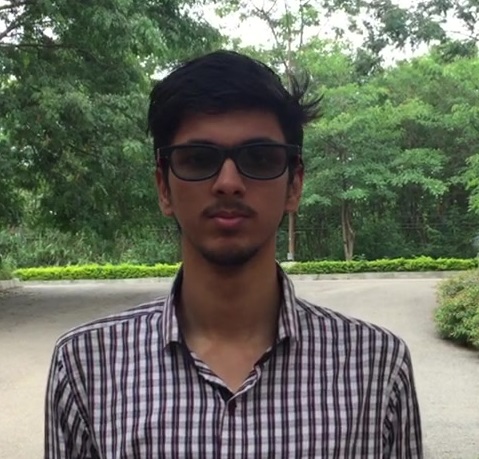} }}
    
        \quad
        \subfloat{{\includegraphics[width=3cm]{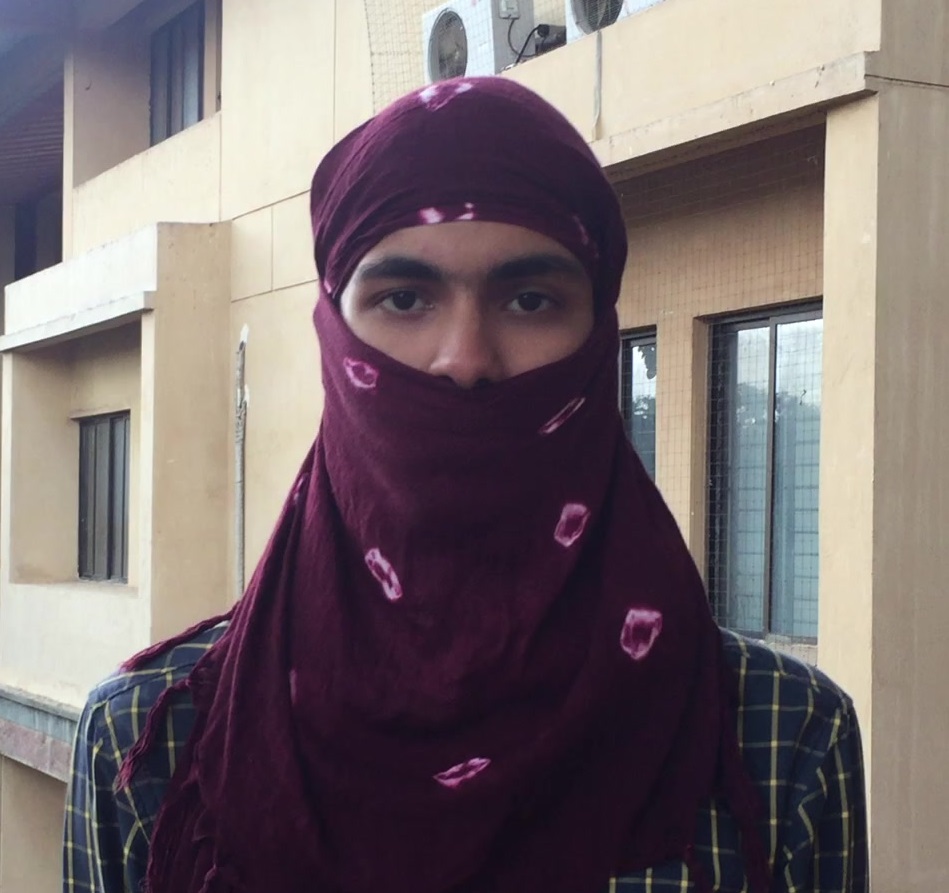} }}
    \subfloat{{\includegraphics[width=3cm]{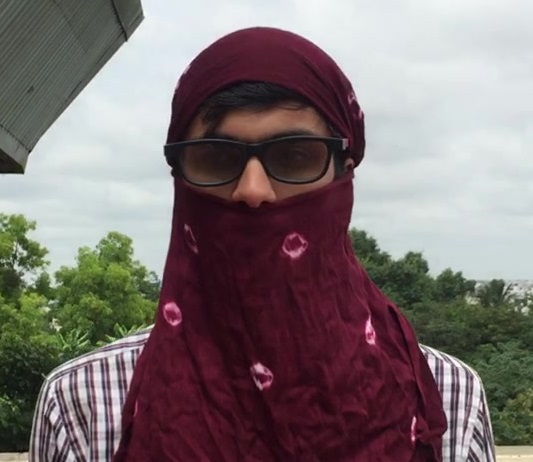} }}
    
    \caption{The backgrounds are complex and semi-complex as seen above. Even with the disguised faces and dynamic backgrounds, our algorithm performs decently.}
    \label{fig:example}
\end{figure}

Moreover, our dataset contains a total of 4,000 images of 20 different subjects taken under varying illumination conditions. There are 8 different backgrounds and the faces exhibit 5 different viewpoints ranging from -20$^\circ$ to 20$^\circ$ (Fig. 3). As shown in Fig. 2, there are 10 different disguises used in this experimental research which are :
\begin{itemize}
\item Beard,
\item Cap,
\item Glasses,
\item Scarf, 
\item Cap \& Glasses, 
\item Beard \& Glasses, 
\item Beard \& Cap, 
\item Scarf \& Glasses, 
\item Cap \& Scarf, 
\item Cap, Glasses \& Scarf.
\end{itemize}

\begin{figure*}
  \includegraphics[width=\textwidth]{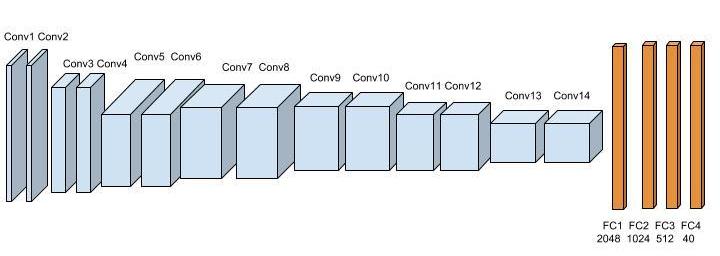}
  \caption{Our DFR Key-Point Prediction Architecture. The large number of convolution layers allow for a greater extent of feature extraction and help improve the accuracy.}
\end{figure*}

\subsection{Preprocessing}

Preprocessing of the image data was carried out by hand annotating each of the 4000 images with 20 facial key-point coordinates. Then each image was resized to 227$\times$227, along with its respective key-point coordinates in the same ratio. Out of 4000 images, 3500 images were used for training our network and 500 images were used for verification. An example of annotation is depicted by Fig. 6

\begin{figure}[h]
\centering
\includegraphics[width=3.1 in]{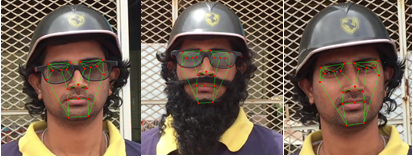}
\caption{Demonstration of Hand-Annotation on Training Dataset}
\label{fig_chi_dot}
\end{figure}

\subsection{The Key-Point Detection Model}

Our proposed framework for facial key-point detection uses a CNN (with 4 fully-connected layers), as shown in Fig. 6 for detecting 20 key-points. This particular detection task is formulated as a regression problem that can be modelled by a CNN. The network takes an image as input and predicts the $(x,y)$ coordinates of learnt facial key-points as output. The network is trained on 3500 images with a batch size of 50 and tested on 500 images. Our architecture efficiently learns the alignment of facial key-points. The predicted coordinates of each key-point from the CNN are overlaid on the respective test images for visual confirmation as could be seen in Fig. 9. The CNN architecture as already explained previously takes images of size 227$\times$227, with a total of 14-convolution layers and 4-fully connected layers as shown in the Fig. 5. The batch size chosen for this experiment was 50 with the \textit{adam} optimizer function, with the system training for 1300 epochs.

\section{Experiments, Evaluation and Results}

As far as our knowledge is concerned, we are not aware of any existing benchmarking techniques for the task of complex disguised facial recognition. In order to overcome that, we have used standard equations for calculating errors in a given dataset, as described in Eq. 1 (where we use Mean Absolute Error) and 2 (where we calculate the Euclidean distance between two spatial points). 

\begin{figure}[h]
\centering
\includegraphics[width=3.2 in]{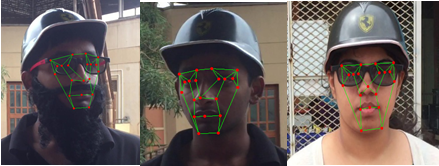}
\caption{A Prediction Example of Our Model on Unseen Images}
\label{fig_chi_dot}
\end{figure}

\subsection{Experimental Setup}
We have used Keras with Tensorflow [17] as the deep learning platform for this experiment. Considering the computational complexity of CNNs and the large amount of time required for training, we trained our model on a system with the configuration as mentioned in Table I.

\captionof{table}{Training System Specifications} \label{tab:title} 
\centering
\begin{tabular}{cccc}
  \hline
  \textbf{Hardware} & \textbf{Specification}\\
  \hline
Memory & 32 GB\\
Processor & Intel Core i7-4770 CPU $@$ 3.4 GHz x 8\\
GPU & GeForce GTX 1050 Ti/PCIe/SSE2\\
OS Type & 64-bit Ubuntu 16.04 LTS\\
  \hline
  \end{tabular}%  <-- note the "%" symbol
\justify

\subsection{Key-point Detection Performance}
We present the performance analysis for detecting each facial key-point. The average key-point detection accuracy of our CNN framework is 65\%. The loss function used in detecting facial key-points is \textit{Mean Absolute Error} (MAE) defined in the following equation :

\begin{equation}
MAE = \frac{1}{n}\sum_{i = 1}^n |\phi_i - \hat{\phi}|
\end{equation}
 
$\phi_i$ represents the position of $i^{th}$ facial key-point of annotated image and $\hat{\phi}$ represents the predicted position of the $i^{th}$ facial key-point of the corresponding image.
Fig. 8 represents graphically the average error in detecting each facial key-point with respect to the ground truth values. 
  
\begin{figure}[h]
\centering
\includegraphics[width=3.4 in]{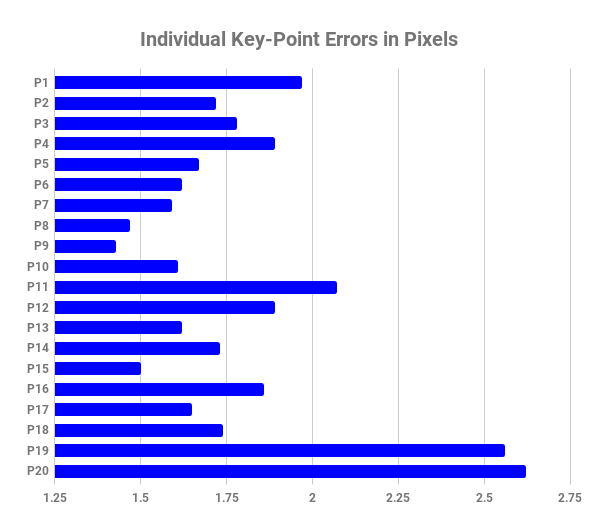}
\caption{Depiction of the Average Error of Prediction of every individual key-point (Y-Axis). As it is clearly visible, the maximum variation of the prediction is by 2-3 spatial units (X-Axis) only. The difference between the prediction and the ground truth is measured using the standard Euclidean Distance Formula. In order to calculate this error metric, the testing (unseen) images were first hand annotated by experts. These images were later fed to model for prediction. The metric suggests how much was a particular predicted key-point is offset from the actual.}
\label{fig_chi_dot}
\end{figure}

Standard Euclidean Distance Formula was used to calculate the error between the predicted key-points $x_1, y_1$ and annotated key-points $x_2, y_2$ :

\begin{equation}
D = \sqrt{(x_1 - x_2)^2 + (y_1 - y_2)^2}
\end{equation}

\subsection{Performance of the Classification Model : The SVM}

As suggested by Philips \textit{et. al.} [18], Guo \textit{et. al.} [19] and Deniz \textit{et. al.} [20], support vector machines have been used extensively for facial recognition and face classification problems. Moreover, it is common knowledge that SVM-based classifications are usually very efficient computationally, so the above factors inspired us to use support vector machines for key-points based face disambiguation. The classification model utilize the ratios and orientations of various connected facial key-points, as shown in Fig. 11, to perform classification task. Our SVM classifier with default parameters classifies the purely complex background disguised faces with an average classification accuracy of 72.4\%.

\begin{figure}[h]
\hspace*{-0.6cm}
\centering
\includegraphics[width=3.6 in]{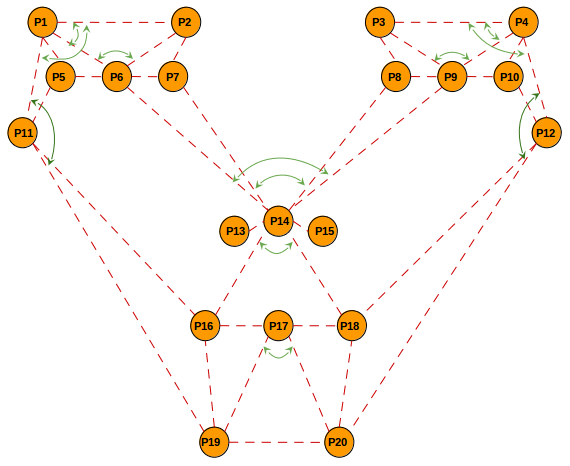}
\caption{A graphical visualization of the method of calculating relative angles and ratios. }
\label{fig_chi_dot}
\end{figure}

The angles marked with green are taken into consideration because it was observed that only the selected angles show considerable variation in the different subjects. Firstly, the slope of the lines is calculated using 

\begin{equation}
m = \tan \theta = \frac{y_2 - y_1}{x_2 - x_1}
\end{equation}

Then the angles between the lines is found using :

\begin{equation}
\phi = \tan^{-1} \frac{m_1 - m_2}{1 + m_1m_2}
\end{equation}
 The results are tabulated graphically in Fig. 12.
\begin{figure}[h]
\hspace*{-0.6cm}
\centering
\includegraphics[width=3.6 in]{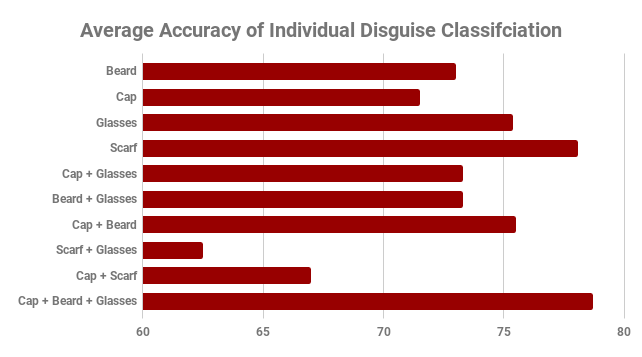}
\caption{Figure shows the classification results of our SVM classifier. It is interesting to note that the accuracy drops between 60-68\% when the disguise contains a scarf. That happens primarily because of the fact that the nose key-point is hidden in these disguises which plays a crucial role in disguised face recognition. Rest all disguises where the point is visible the classification rates are higher. The results have been evaluated on 500 unseen images and presented as the average of the classification rates.}
\label{fig_chi_dot}
\end{figure}

\subsection{Time Complexity}

We have used the Python's \textit{timeit} library to evaluate the computational complexity of our prediction. We have observed that it takes 2.598 seconds to process a batch of 50 frames, which means that it takes around 0.0518 seconds/frame, which eventually results in a decent FPS rate of 19.3 frames/second.

\subsection{Hardware Implementation}

As of now, the algorithm has been demonstrated on a portable system with the configuration as mentioned in Table II. We propose to use our suggested method in security-demanding regions where the minimum configuration of the systems are similar to the one in Table II. The camera orientation plays a big role in this as the subject needs to be in the plane of view for active recognition.

\captionof{table}{Real-Time Test Bench Specifications} \label{tab:title} 
\centering
\begin{tabular}{cccc}
  \hline
  \textbf{Hardware} & \textbf{Specification}\\
  \hline
Memory & 8 GB\\
Processor & Intel Core i5-4770 CPU $@$ 2.3 GHz x 2\\
GPU & N/A\\
OS Type & 64-bit Windows 7 Pro\\
  \hline
  \end{tabular}%  <-- note the "%" symbol
\justify
%\begin{figure}[h]
%\centering
%\includegraphics[width=2.5 in]{1}
%\caption{Parrot Bebop 2}
%\label{fig_chi_dot}
%\end{figure}

\subsection{Comparison with the State-of-the-Art}

Since the dataset used in this research is same as that used by Singh \textit{et. al.} [16], we present similar metrics. The classification accuracy of our SVM outperforms the existing state-of-the-art methodology by a margin of 9.8\%. We report the classification rates on both simple and complex dataset, as tabulated in Table III. As it would have been observed, not much attention was given to simple dataset classification because the simple dataset consists of disguised faces without any complex background which is usually not the real-life scenario always..

\captionof{table}{Performance Results} \label{tab:title} 
\centering
\hspace*{-0.4cm}
\begin{tabular}{cccc}
  \hline
  \textbf{Algorithms} & \textbf{Simple Dataset} & \textbf{Complex Dataset}\\
  \hline
Dhamecha \textit{et. al.} [2] & 65.2\%      & 53.4\%       \\
Singh \textit{et. al.} [3] & 78.4\%       & 62.6\%     \\
 DFR (\textit{Ours})  & \textbf{86.6\%}       & \textbf{72.4\%}     \\
  \hline
  \end{tabular}%  <-- note the "%" symbol

\justify

%\captionof{table}{Classification Rates of Various Algorithms} \label{tab:title} 
%\begin{tabular}{ |l|l|l| }
%%\hline
%%\multicolumn{3}{ |c| }{Comparison Chart} \\
%\hline
%%Goalkeeper & GK & Paul Robinson \\ \hline
%\multirow{3}{*}{Dhamecha \textit{et. al.} [2]} & Simple Dataset & 65.2 \\
% & Complex Dataset & 53.4 \\ \hline
%\multirow{2}{*}{Singh \textit{et. al.} [3]} & Simple Dataset & 78.4 \\
% & Complex Dataset & 62.6 \\ \hline
%%Forward & FW & Jamie McMaster \\ \hline
%\multirow{1}{*}{DFR (\textit{Ours})} & Simple Dataset &  \textbf{86.6} \\
% & Complex Dataset & \textbf{72.4} \\ \hline
%\hline
%\end{tabular}

\justify

\section{Supplementary Material}
All the codes written during the conducted research (training, testing and data annotation) along with the trained models have been made available at the following GitHub repository : \href{https://github.com/abrarmajeedi/Disguised-Facial-Recognition-DFR.git}{https://github.com/abrarmajeedi/Disguised-Facial-Recognition-DFR.git}

\section{Future Work and Conclusion}
%\addtolength{\textheight}{-12cm}   % This command serves to balance the column lengths

The problem of identifying masked faces is a challenge difficult enough for human beings themselves. So, imparting that amount of intelligence to a machine would take time and training. Moreover, even though there have been tremendous advancements in the field of CNNs, a lot still remains to explore. The algorithm presented in this paper attempts to utilize the beauty of these networks and support vector machines for aforementioned problem statement of identifying disguised faces.. Although the classification rate is not totally inaccurate, there is a great scope for further improvement. One aspect would be the construction of an extensive dataset for disguised faces in the wild. One that would multiple orientations of the faces in multiple spectrum apart from RGB like NIR etc. Probably a larger neural network might also help improve the key-point detection rates, but that might also affect the real-time performance of the system.

Another sub-section of further research constitutes the camera positioning, the active vision part. The camera may or may not be facing the suspect directly which could result in some degradation in the performance. The algorithm could be implemented on a regular PTZ (pan-tilt-zoom) camera with ethernet-based video transmission system. Such cameras are usually placed in strategic locations in high-security zones. With the different controls available with the hardware, the user could directly focus the camera onto the incoming people, which could make face description easier. We aim to improve upon the above mentioned research components in the next stages of this project. However, the major contribution of this paper lies in the application of the DFR model in real-time, which could probably lead to a concrete security system design someday.

\end{document}